%% file: coling20-quality-assessment-frame.tex
\documentclass[11pt]{article}
\usepackage{coling2020}
\usepackage{times}
\usepackage{url}
\usepackage{latexsym}

\usepackage[T1]{fontenc}
\usepackage{type1cm}
\usepackage[american]{babel}
\usepackage{url}
\usepackage{amssymb} 
\usepackage{amsmath}
\usepackage{multirow} 
\usepackage{xspace}
\DeclareMathAlphabet{\mathcalbf}{OMS}{pzc}{b}{n}
\usepackage{microtype}

\usepackage{tabularx}
\usepackage{booktabs}
\usepackage{graphicx}
\DeclareGraphicsExtensions{.pdf,.ai,.jpg,.png}
\setkeys{Gin}{pagebox=artbox}% Use artbox instead of mediabox default. Options: mediabox, cropbox, bleedbox, trimbox, artBox. (shell: pdfinfo -box <pdf-file>)

\graphicspath{{../coling20-quality-assessment-figures/}}

\newcommand{\bsfigure}[3][]{%
	\begin{figure}[t]
		\centering
		\includegraphics[#1]{#2}
		\vglue -1ex plus 0.0ex minus 0.5ex
		\caption{#3}\label{#2}%
 	 \end{figure}
}

\RequirePackage{type1cm}% CM-Fonts including Math-Fonts in all sizes.
\RequirePackage{color}
\RequirePackage{soul}
\setstcolor{blue}

\newsavebox\bscombox
\newcommand{\bscom}[3][]{%
  \sbox{\bscombox}{\fontsize{8}{9}\selectfont#1#2#3}
  \noindent
  \st{#2}{\selectfont
    \color{blue}#3\ifx\\#1\\\else{\fontsize{8}{9}\selectfont\color{violet}[#1]}\fi
    }
  }

\definecolor{highlight1}{rgb}{0.95,0.95,0.95}
\definecolor{tgray}{rgb}{0.5,0.5,0.5}
\newcommand{\gr}{\color{tgray}}
\newcommand{\bl}{}

\begin{document}

%
% The following footnote without marker is needed for the camera-ready
% version of the paper.
% Comment out the instructions (first text) and uncomment the 8 lines
% under "final paper" for your variant of English.
% 
\blfootnote{
    %
    % for review submission
    %
%    \hspace{-0.65cm}  % space normally used by the marker
%    Place licence statement here for the camera-ready version. 
    %
    % % final paper: en-us version 
    %
    \hspace{-0.65cm}  % space normally used by the marker
    This work is licensed under a Creative Commons 
    Attribution 4.0 International License.
    License details:
    \url{http://creativecommons.org/licenses/by/4.0/}.
}

\input{coling20-quality-assessment-pre}
\input{coling20-quality-assessment-part1}

\input{coling20-quality-assessment-part2}
\input{coling20-quality-assessment-part3}

\input{coling20-quality-assessment-part4}

\input{coling20-quality-assessment-part5}

\input{coling20-quality-assessment-sum}

\bibliographystyle{acl}
\bibliography{coling2020}

\end{document}

%% file: coling20-quality-assessment-pre.tex
\title{Intrinsic Quality Assessment of Arguments}

\author{Henning Wachsmuth \\
  Department of Computer Science \\
  Paderborn University \\
  Paderborn, Germany \\
  {\tt henningw@upb.de} \\\And
Till Werner \\
  Department of Computer Science \\
  Paderborn University \\
  Paderborn, Germany \\
  {\tt wtill@mail.upb.de} \\}

\date{}

\maketitle

\begin{abstract}
Several quality dimensions of natural language arguments have been investigated. Some are likely to be reflected in linguistic features (e.g., an argument's arrangement), whereas others depend on context (e.g., relevance) or topic knowledge (e.g., acceptability). In this paper, we study the {\em intrinsic} computational assessment of 15 dimensions, i.e., only learning from an argument's text. In systematic experiments with eight feature types on an existing corpus, we observe moderate but significant learning success for most dimensions. Rhetorical quality seems hardest to assess, and subjectivity features turn out strong, although length bias in the corpus impedes full validity. We also find that human assessors differ more clearly to each other than to our approach.
\end{abstract}

%% file: coling20-quality-assessment-part1.tex
\section{Introduction}
\label{sec:introduction}

Good arguments help to persuade people, to compromise, or to at least understand each other better. What quality dimension is meant by {\em good} depends on the setting, though \cite{vaneemeren:2004,johnson:2006}. Several dimensions may be assessed computationally, as we exemplify for an argument in favor of advancing the common good, taken from \newcite{wachsmuth:2017d}:
\begin{quote}
\em
``While striving to make advancements for the common good you can change the world forever. Allot of people have succeded in doing so. Our founding fathers, Thomas Edison, George Washington, Martin Luther King jr, and many more. These people made huge advances for the common good and they are honored for it.''
\end{quote}

\noindent
The argument is {\em well-organized} \cite{persing:2010}, its premises are certainly largely {\em acceptable} \cite{yang:2019} and {\em relevant} to the topic \cite{wachsmuth:2017a}. Whether they {\em suffice} to draw the conclusion \cite{stab:2017} is another question, let alone how {\em convincing} the argument is \cite{habernal:2016a}. Some dimensions may be reflected in linguistic features of an argument's text. Others depend on context, require topic or background knowledge, or are inherently subjective.

In this paper, we benchmark what quality dimensions of an argument can be assessed intrinsically, i.e., when analyzing the text of an argument only. Given the corpus of \newcite{wachsmuth:2017b} with 304 English debate portal arguments on 16 topics scored for 15 logical, rhetorical, and dialectical quality dimensions by three experts, we carry out systematic leave-one-topic-out cross-validation experiments. In particular, we learn standard supervised score regression on various text features; from content and distributional semantics, to style, structure, and length, to text quality, evidence, and subjectivity. 

For 11 dimensions, we observe moderate but significant prediction gains of the features over a mean baseline. Following intuition, rhetorical quality related to credibility and emotions seem hardest to assess. Features capturing subjectivity (e.g., sentiment and pronoun usage) turn out particularly effective. Even better performs the length feature, though, revealing bias in the corpus and matching previous findings on other corpora \cite{potash:2017}. Follow-up experiments indicate that the experts strongly differ in their assessability, and that some beat our features only slightly in assessing quality. Altogether, an intrinsic computational assessment of argument quality seems useful but not enough alone.\,\,\,

%% file: coling20-quality-assessment-part2.tex
\section{Related Work}
\label{sec:relatedwork}

Soon after the rise of argument mining, argument quality assessment has come up as a task \cite{stede:2018}, due to its importance for applications such as {\em Project Debater} \nopagebreak{\cite{gleize:2019}.} It rests on extensive theoretical discussions about what good arguments \cite{johnson:2006} and bad arguments are \cite{walton:2006}, and how to argue reasonably \cite{vaneemeren:2004}.

% NOTE. Left out: stab:2016, simpson:2018, gleize:2020
Several corpora and approaches were proposed for specific argument quality dimensions, first related to essay scoring \cite{persing:2015}, some of which  modeling arguments explicitly \cite{wachsmuth:2016}. Later approaches targeted arguments from debate portals \cite{wei:2016}, student essays \cite{stab:2017}, mixed web texts \cite{wachsmuth:2017a}, and news editorials \cite{yang:2019,elbaff:2020a}. We use the corpus of \newcite{wachsmuth:2017b}, as it is the only one annotated for diverse dimensions and is claimed to reflect argument quality comprehensively. In follow-up work, \newcite{wachsmuth:2017d} found correlations with the convincingness reasons of \newcite{habernal:2016b}, and \newcite{potthast:2019} as well as \newcite{gretz:2020} have evaluated their annotations against the quality annotation scheme of the corpus. However, we do not know any previous assessment approach developed on the corpus, possibly due to its limited size (see Section~\ref{sec:data}). 

We focus on features intrinsic to an argument's text. This complements the study of \newcite{potash:2017} who employed external knowledge to assess convincingness. Like them, we find that longer arguments tend to be judged better. \newcite{toledo:2019} limit arguments to at most 36 words, avoiding length bias but also preventing deeper reasoning. Quality in their corpus reflects which argument is in doubt preferred. Ultimately, such judgments remain subjective \cite{lukin:2017}. To alleviate this, \newcite{elbaff:2018} encode the reader's ideology and personality, but such information is often not given in practice.

%% file: coling20-quality-assessment-part3.tex
\section{Data}
\label{sec:data}

The corpus of \newcite{wachsmuth:2017b} is a subset of 320 debate portal posts from the dataset of \newcite{habernal:2016b}, 20 each for 16 controversial topics. Three human experts (all authors of the paper) scored all posts that they saw as arguments for the following 15 logical, rhetorical, and dialectical quality dimensions on a scale from 1~(low) to 3~(high). In line with the experiments of \newcite{wachsmuth:2017d}, we use only those 304 texts in Section~\ref{sec:experiments} that were seen as arguments by all three experts. 

\paragraph{Logic}

The main logical dimension is called {\em cogency (Cog)}. It is defined based on three subdimensions: the {\em local acceptability (LAc)} of the truth of an argument's premises, the premises' {\em local relevance (LRe)} for the argument's conclusion, and their {\em local sufficiency (LSu)} to infer the conclusion.

\paragraph{Rhetoric}

The main rhetorical dimension is the {\em effectiveness (Eff)} in persuading readers. Subdimensions are the argument's {\em clarity (Cla)}, the author's {\em credibility (Cre)}, the {\em appropriateness (App)} of the argument's used language, its success in {\em emotional appeal (Emo)}, and its sequential {\em arrangement (Arr)} in the text.

\paragraph{Dialectic}

The main dialectical dimension is {\em reasonableness (Rea)}, with three subdimensions: the {\em global acceptability (GAc)} of stating the argument when discussing a given issue, the argument's {\em global relevance (GRe)} for achieving agreement, and its {\em global sufficiency (GSu)} in discussing both sides of the issue.

\paragraph{Overall}
The {\em overall quality (OvQ)} reflects the subjective weighting of all 14 other quality dimensions.

\medskip
\noindent
Both the single expert scores and the mean score are provided for each dimension. The inter-annotator agreement for the different dimensions ranged from 0.26 (emotional appeal) to 0.51 (overall quality) in terms of Krippendorff's~$\alpha$. The majority score of most dimensions is~2, but both score~1 and~3 also occur frequently for many of them. Matching the hierarchical idea of the dimensions, overall quality correlates strongest with cogency, effectiveness, and reasonableness. For details on agreements, score distributions, and correlations, we refer the reader to the original paper \cite{wachsmuth:2017b}.

%% file: coling20-quality-assessment-part4.tex
\section{Approach}
\label{sec:approach}

This paper does {\em not} aim to propose a novel quality assessment approach, but to evaluate what features of a text help to assess which quality dimension. External knowledge is included only via lexicons and embeddings. As an example, Figure~\ref{assessment} shows selected textual aspects of the argument from Section~\ref{sec:introduction} that may be predictive of certain dimensions. We quantify these and other aspects in the following eight feature types that are employed in linear SVMs for score regression \cite{chang:2011}:%
\footnote{We also tested some pretrained configurations of BERT \cite{devlin:2018}, fine-tuning them during training. However, performance was low, possibly due to the small data and the prevention of learning topic information in our experimental setup.}

\bsfigure{assessment}{Exemplary analysis  of an argument from the used corpus for selected text features that might affect quality: content-related {\em key phrases}, text quality indicators such as {\em spelling errors}, subjective {\em pronoun usage}, length in {\em sentences/tokens}, and evidence distributions reflected by {\em premises} and {\em conclusions}. On the right, all mean quality scores of the argument in the corpus are shown (worst is 1, best is 3).}

\paragraph{Content}

As often done in text classification \cite{aggarwal:2012}, we aim to capture important content-related key phrases simply as part of the distribution of word 1- to 3-grams, taking all those that occur in $\geq$ 3\% of all training texts (such thresholds were set after initial tests).

\paragraph{Embedding}

We capture an argument's distributional semantics by a sentence vector, using the pretrained {\em fasttext} model based on Wikipedia \cite{mikolov:2018}. Each vector position becomes one feature.

\paragraph{Style}

We model style with part-of-speech 1- to 3-grams ($\geq$10\% training frequency) and character 1- to 3-grams ($\geq$3\%). Both are common stylometric features in authorship attribution \cite{stamatatos:2009}.

\paragraph{Structure}

In terms of structure, we look for enumeration indicators, such as ``1.'' and ``2.'' or ``on one hand'' and ``on the other hand''. In addition, we check the first token 1-, 2-, and 3-gram in the text.

\paragraph{Length}

Our length feature type includes normalized counts of characters, syllables, tokens, phrases, sentences, and paragraphs, along with ratios between each pair of these linguistic units.

\paragraph{Text Quality}

 Motivated by classical essay scoring \cite{ke:2019}, we model text quality by spelling correctness and readability. The former is quantified as absolute and relative counts of three error types from {\em www.languagetool.org} (hints, unknown words, and others). For the latter, we calculate 10 common readability scores, including Flesch Kincaid Reading Ease, Gunning Fog Index, LIX, and similar.

\paragraph{Evidence}

On one hand, we capture the evidence given in an argument in terms of the frequency of links. On the other hand, we apply an out-of-the-box argument mining algorithm \cite{wachsmuth:2016} that classifies each sentence into one of four argumentative unit types: thesis, conclusion, premise, or none.

\paragraph{Subjectivity}

Since we hypothesized subjectivity to be important, we here combine multiple frequency distributions: (a) singular and plural 1st, 2nd, and 3rd person pronouns, (b) indicators of positivity, negativity, and hedging (e.g., ``certainly'') based on \newcite{rittman:2004}, (c) 83 different emojis, (d) fully lower-case, upper-case, and other words, and (e) character types, such as letter, digit, and whitespace.

% NOTE. Left out.
%\subsection{BERT-based Classification of Argument Quality Scores}
%
%In our final experiment, we also include a simple pre-trained BERT model to predict score 1, 2, or 3 for each argument. We fine-tune the model on the respective training set in each run, using \bscom{}{xxx} as the learning rate and \bscom{}{standard values} for all other parameters.

%% file: coling20-quality-assessment-part5.tex
\section{Experiments}
\label{sec:experiments}

We now report on experiments with the features from Section~\ref{sec:approach} on the corpus from Section~\ref{sec:data}. In particular, we systematically study three research questions for the 15 given argument quality dimensions:
\begin{enumerate}
\setlength{\itemsep}{0pt}
\item[Q1.]
To what extent can each quality dimension be assessed only from an argument's text?
\item[Q2.]
How dependent is the assessability on the subjective view of the experts?
\item[Q3.]
How well do the considered features predict argument quality compared to humans?
\end{enumerate}

\paragraph{Experimental Setup}

We approached all 15 dimensions using each feature type alone, feature ablation (all but one type), and all features respectively. We split the corpus into 16 test sets, one per topic. For each approach and topic, we trained one SVM on the other 15 topics, optimizing its $C$ hyperparameter in 15-fold cross-validation on the training set (tested $C$ range: $10^{-4} \cdot 2^{j}$ for $7 \leq j \leq 16$). %Given that no big outliers exist for the quality score range (1--3), 
We then computed the {\em mean absolute error (MAE)}, averaged over the 16 MAEs on each test set. This leave-one-topic-out way ensures that no topic information can be exploited in the assessment on the test sets.

To focus on the learning success, we compare the features only to the {\em mean baseline}, which always predicts the mean score of all arguments in the given training set. For the SVM with all features, we use the 16 single MAEs in a one-tailed independent $t$-test  to test whether improvements over the baseline are significantly better at $p < .05$ (marked $\dagger$ below) and $p < .01$ ($\ddagger$). The Java code for reproducing the experiments can be accessed here: \url{http://arguana.com/software}

\input{table-features}

\paragraph{Quality Assessment (Q1)}

To provide answers to question~Q1, we let all SVMs learn to assess the mean score of the three experts. Table~\ref{table-features} shows the MAE of each feature type ($A_i$), feature ablation ($A_{\backslash i}$), and all features ($A_{1-8}$) in comparison to the baseline for each quality dimension. The SVM with all features ($A_{1-8}$) outperforms the baseline in all cases. Only for four dimensions, the gains are not significant, three of which being rhetorical: {clarity~({\em Cla}), credibility ({\em Cre}), and emotional appeal ({\em Emo}). This may be due to their subjective nature, as reflected in limited inter-annotator agreement \cite{wachsmuth:2017b}. The highest MAE reduction is achieved for local sufficiency (0.39 to 0.30), which has also been successfully assessed in previous studies \nopagebreak{\cite{stab:2017}}. Other clear gains are achieved for overall quality (0.45 to 0.37) and for cogency (0.44 to 0.37). No dimension is really ``solved'' by the given features, but we conclude that intrinsic argument quality assessment is effective to some extent. 

Looking at the features, we see that {\em content ($A_1$)} and {\em embedding ($A_2$)} perform rather badly, unlike in many NLP tasks. This is somewhat expected, though, due to our leave-one-topic-out setting. While feature ablation leads to the best results for some dimensions (e.g., {\em Cre} and {\em Arr}), two feature types dominate the assessment:  {\em Subjectivity ($A_8$)} alone minimizes the~MSE for five dimensions, once being the single best approach (for local relevance, {\em LRe}). However,  {\em length ($A_5$)} is even stronger, e.g., being best for reasonableness and overall quality. Albeit quality may require some words, this reveals length bias inherent to the corpus. Such bias was also found in other argument quality corpora \cite{potash:2017}. It questions the validity of the annotated scores, even if $A_5$ is not needed for many dimensions (see $A_{\backslash 5}$).

\input{table-subjectiveness}

\paragraph{Subjectiveness (Q2)}

For Q2, we learn to assess the score of each single expert and compare our features to the baseline in Table~\ref{table-subjectiveness}. {\em Mean scores} lead to the lowest MAE, due to their natural tendency towards middle scores. We find clear differences between the experts, reflecting how subjective the assessment is: Hardly any significant learning success is observed on the scores of {\em expert \#2}, whereas particularly~{\em \#3} seems well-assessable. While this may mean that some experts are either more reliable or more influenced by surface text features, it raises the question whether assessing the mean score is the best choice.

\paragraph{Human vs.\ Machine (Q3)}

For Q3, finally, we evaluate how much the experts diverge from the majority score as opposed to the all-features SVM. Since the experts could only give integer scores, for fairness we rounded the scores of the SVM before MAE computation. Still, Table~\ref{table-human-machine} reveals that {\em expert \#2} significantly beats our features only on three dimensions ({\em Eff}, {\em GAcc}, and {\em OvQ}) and is even worse on five dimensions ({\em App}, {\em Emo}, {\em Arr}, {\em GRe}, and {\em GSu}). So, our features can compete with some humans. {\em Expert \#3}, in contrast, clearly outperforms the SVM with a very low MAE for most dimensions. Together with the results on Q2, it seems that some expert scores are more consistent.\,\,

\input{table-human-machine}

%% file: table-features.tex
\begin{table*}[t!]
\small
\centering
\setlength{\tabcolsep}{2pt}
\begin{tabular*}{\linewidth}{ll@{}crrrrcrrrrrrcrrrrcr}
\toprule																				
		&						&	& \multicolumn{4}{c}{\bf Logical quality}	&	& \multicolumn{6}{c}{\bf Rhetorical quality}					&	& \multicolumn{4}{c}{\bf Dialectical quality}			\\		
									\cmidrule(l@{2pt}r@{2pt}){4-7}				\cmidrule(l@{2pt}r@{2pt}){9-14}							\cmidrule(l@{2pt}r@{2pt}){16-19}									 	
\bf \#		& \bf Approach				&	& \bf Cog	& \bf LAc	& \bf LRe	& \bf LSu	&	& \bf Eff	& \bf Cla	& \bf Cre	& \bf App	& \bf Emo	& \bf Arr	& 	& \bf Rea	& \bf GAc	& \bf GRe	& \bf GSu	& 	& \bf OvQ	\\
\midrule
$A_{1}$	& Content					&	& 0.38	& 0.42	& 0.43	& 0.32	& 	& 0.34	& 0.38	& 0.31	& 0.36	& 0.30	& 0.36	& 	& 0.39	& 0.41	& 0.40	& 0.24	& 	& 0.39	\\
$A_{2}$	& Embedding  				&	& 0.46	& 0.44	& 0.48	& 0.38	& 	& 0.39	& 0.38	& 0.32	& 0.36	& \bf 0.28	& 0.39	& 	& 0.45	& 0.47	& 0.44	& 0.28	& 	& 0.45 	\\
$A_{3}$	& Style					&	& 0.38	& 0.41	& 0.44	& 0.31	& 	& \bf 0.33	& 0.39	& 0.30	& 0.36	& 0.29	& 0.36	& 	& 0.38	& 0.40	& 0.38	& 0.23	& 	& 0.38	\\
$A_{4}$	& Structure				&	& 0.45	& 0.46	& 0.49	& 0.38	& 	& 0.39	& 0.39	& 0.34	& 0.38	& 0.30	& 0.41	& 	& 0.46	& 0.46	& 0.44	& 0.28	& 	& 0.46	\\
$A_{5}$	& Length					&	& 0.37	& \bf 0.40	& 0.43	& \bf 0.30	& 	& \bf 0.33	& 0.38	& 0.30	& 0.34	& \bf 0.28	& 0.35	& 	& \bf 0.36	& \bf 0.38	& \bf 0.37	& \bf 0.22	& 	& \bf 0.36	\\
$A_{6}$	& Text quality				&	& 0.44	& 0.45	& 0.48	& 0.37	& 	& 0.38	& 0.38	& 0.33	& 0.37	& 0.31	& 0.39	& 	& 0.45	& 0.44	& 0.44	& 0.27	& 	& 0.44	\\
$A_{7}$	& Evidence				&	& 0.42	& 0.44	& 0.46	& 0.35	& 	& 0.37	& 0.40	& 0.33	& 0.38	& 0.30	& 0.38	& 	& 0.42	& 0.45	& 0.40	& 0.24	&	& 0.42 	\\
$A_{8}$	& Subjectivity				&	& \bf 0.36	& \bf 0.40	& \bf 0.41	& 0.31	& 	& \bf 0.33	& 0.39	& 0.33	& 0.35	& 0.29	& 0.36	& 	& 0.37	& 0.40	& 0.38	& \bf 0.22	& 	& 0.37	\\
\addlinespace
$A_{\backslash 1}$	& w/o Content		&	& 0.37	& \bf 0.40	& 0.43	& \bf 0.30	& 	& \bf 0.33	& \bf 0.36	& 0.30	& \bf 0.33	& 0.29	& 0.35	& 	& 0.37	& 0.40	& \bf 0.37	& \bf 0.22	& 	& 0.37	\\
$A_{\backslash 2}$	& w/o Embedding	&	& 0.37	& \bf 0.40	& 0.42	& \bf 0.30	& 	& \bf 0.33	& \bf 0.36	& 0.30	& \bf 0.33	& 0.29	& 0.35	& 	& 0.37	& 0.40	& \bf 0.37	& \bf 0.22	& 	& 0.37 	\\
$A_{\backslash 3}$	& w/o Style		&	& \bf 0.36	& \bf 0.40	& 0.42	& \bf 0.30	& 	& \bf 0.33	& \bf 0.36	& 0.31	& \bf 0.33	& 0.29	& \bf 0.34	& 	& 0.37	& 0.39	& \bf 0.37	& \bf 0.22	& 	& 0.37	\\
$A_{\backslash 4}$	& w/o Structure		&	& \bf 0.36	& \bf 0.40	& 0.42	& \bf 0.30	& 	& \bf 0.33	& 0.37	& \bf 0.29	& \bf 0.33	& 0.29	& \bf 0.34	& 	& 0.37	& 0.39	& \bf 0.37	& \bf 0.22	& 	& 0.37	\\
$A_{\backslash 5}$	& w/o Length		&	& \bf 0.36	& 0.41	& 0.42	& \bf 0.30	& 	& \bf 0.33	& \bf 0.36	& 0.30	& \bf 0.33	& \bf 0.28	& \bf 0.34	& 	& 0.37	& 0.40	& \bf 0.37	& \bf 0.22	& 	& 0.37	\\
$A_{\backslash 6}$	& w/o Text quality	&	& 0.37	& \bf 0.40	& 0.42	& \bf 0.30	&	& \bf 0.33	& 0.37	& 0.30	& 0.34	& \bf 0.28	& 0.35	& 	& 0.37	& 0.40	& \bf 0.37	& \bf 0.22	& 	& 0.37	\\
$A_{\backslash 7}$	& w/o Evidence		&	& \bf 0.36	& \bf 0.40	& 0.42	& \bf 0.30	& 	& \bf 0.33	& \bf 0.36	& 0.30	& \bf 0.33	& \bf 0.28	& 0.35	& 	& 0.38	& 0.40	& \bf 0.37	& \bf 0.22	& 	& 0.37	\\
$A_{\backslash 8}$	& w/o Subjectivity	&	& 0.37	& 0.41	& 0.43	& 0.31	& 	& \bf 0.33	& 0.37	& 0.30	& 0.34	& 0.29	& 0.35	& 	& 0.37	& 0.40	& \bf 0.37	& \bf 0.22	& 	& 0.37	\\
\addlinespace
$A_{1-8}$	& All features		&	& $^\ddagger$0.37	& \bf $^\ddagger$0.40	& \phantom{$^\dagger$}0.42	& \bf $^\ddagger$0.30	& 	& \bf $^\ddagger$0.33	& \bf \phantom{$^\dagger$}0.36	& \phantom{$^\dagger$}0.30	& \bf $^\dagger$0.33	& \phantom{$^\dagger$}0.29	& $^\dagger$0.35	&	& $^\dagger$0.37	& $^\dagger$0.40	& \bf $^\dagger$0.37	& \bf $^\dagger$0.22	&	& $^\ddagger$0.37  	\\
\midrule
$B$		& Baseline	 			&	& 0.44	& 0.46	& 0.47	& 0.39	& 	& 0.39	& 0.40	& 0.33	& 0.39	& 0.31	& 0.40	& 	& 0.43	& 0.46	& 0.43	& 0.26	& 	& 0.45  	\\
\bottomrule
\end{tabular*}
\caption{Q1. Mean absolute error of each feature type, feature ablation, all features, and the mean baseline for all 15 quality dimensions, averaged over all 16 test sets. The best value in a column is bold. For {\em all features}, significant improvements over the mean baseline are marked with $\dagger$ ($p\!<\!.05$) and $\ddagger$ ($p\!< \!.01$).} 
\label{table-features}
\end{table*}

%% file: table-subjectiveness.tex
\begin{table*}[t!]
\small
\renewcommand{\arraystretch}{0.96}
\centering
\setlength{\tabcolsep}{1.5pt}
\begin{tabular*}{\linewidth}{ll@{}crrrr@{$\;\;\;$}crrrrrr@{$\;\;\;$}crrrr@{$\;\;\;$}cr}
\toprule										
			&				&	& \multicolumn{4}{c}{\bf Logical quality}	&	& \multicolumn{6}{c}{\bf Rhetorical quality}					&	& \multicolumn{4}{c}{\bf Dialectical quality}			\\		
								\cmidrule(l@{2pt}r@{2pt}){4-7}				\cmidrule(l@{2pt}r@{2pt}){9-14}							\cmidrule(l@{2pt}r@{2pt}){16-19}	
\bf Scores		& \bf Approach		&	& \bf Cog	& \bf LAc	& \bf LRe	& \bf LSu		&	& \bf Eff	& \bf Cla	& \bf Cre	& \bf App	& \bf Emo	& \bf Arr	& 	& \bf Rea	& \bf GAc	& \bf GRe	& \bf GSu	& 	& \bf OvQ	\\			
\midrule
Expert \#1		& All features		&	& $^\dagger$0.57 	& 0.57	& 0.52	& \bf $^\ddagger$0.47		& 	&  $^\dagger$0.54	& 0.52	&  0.41 	& 0.51	&  0.35	&  0.54 	&	& $^\dagger$0.52	& 0.56	& 0.54	& \bf $^\ddagger$0.45	& 	& \bf $^\ddagger$0.52 	\\
			& Baseline		&	& \bl 0.64	& \bl 0.60	& \bl 0.59	& \bl 0.57		&	& \bl 0.61	& \bl 0.52	& \bl 0.48	& \bl 0.51	& \bl 0.34	& \bl 0.55	&	& \bl 0.59	& \bl 0.59	& \bl 0.56	& \bl 0.54	&	& \bl 0.60	\\[0.5ex]
Expert \#2		& All features		&	& 0.49	& 0.59	& 0.63	& 0.34		& 	& 0.48	& 0.53	& 0.37	& 0.67	& 0.31	& 0.49	& 	& 0.59	& 0.56	& 0.58	& 0.26	& 	& \bf $^\ddagger$0.50	\\
			& Baseline		&	& \bl 0.57	& \bl 0.60	& \bl 0.67	& \bl 0.44		&	& \bl 0.56	& \bl 0.53	& \bl 0.40	& \bl 0.64	& \bl 0.28	& \bl 0.53	&	& \bl 0.62	& \bl 0.56	& \bl 0.63	& \bl 0.30	&	& \bl 0.62	\\[0.5ex]
Expert \#3		& All features		&	& \bf $^\ddagger$0.50	& $^\dagger$0.51	& \bf $^\dagger$0.53	& \bf $^\ddagger$0.47		& 	& $^\dagger$0.38	& 0.54	& \bf $^\dagger$0.48	& 0.33	& 0.44	& 0.50	& 	& \bf $^\ddagger$0.46	& 0.47	& 0.53	& $^\dagger$0.32	& 	& \bf $^\ddagger$0.47	\\
			& Baseline		&	& \bl 0.60	& \bl 0.57	& \bl 0.59	& \bl 0.60		&	& \bl 0.45	& \bl 0.54	& \bl 0.53	& \bl 0.37	& \bl 0.51	& \bl 0.53	&	& \bl 0.55	& \bl 0.51	& \bl 0.53	& \bl 0.40	&	& \bl 0.59	\\
%Mean \#1--3	& All features		&	& 0.52	& 0.55	& 0.56	& 0.43		& 	& 0.47	& 0.53	& 0.42	& 0.50	& 0.37	& 0.51	& 	& 0.53	& 0.53	& 0.55	& 0.34	& 	& 0.50 	\\
%			& Baseline		&	& \bl 0.60	& \bl 0.59	& \bl 0.62	& \bl 0.54		&	& \bl 0.54	& \bl 0.53	& \bl 0.47	& \bl 0.51	& \bl 0.38	& \bl 0.54	&	& \bl 0.59	& \bl 0.55	& \bl 0.57	& \bl 0.42	&	& \bl 0.60	\\
\midrule	
Mean score	& All features		&	& \bf $^\ddagger$0.37	& \bf $^\ddagger$0.40	& \phantom{$^\dagger$}0.42	& \bf $^\ddagger$0.30	& 	& \bf $^\ddagger$0.33	& \phantom{$^\dagger$}0.36	& \phantom{$^\dagger$}0.30	& \bf $^\dagger$0.33	& 0.29	& \bf $^\dagger$0.35	&	& $^\dagger$0.37	& \bf $^\dagger$0.40	& \bf $^\dagger$0.37	& $^\dagger$0.22	&	& \bf $^\ddagger$ 0.37  	\\
			& Baseline	 	&	& 0.44	& 0.46	& 0.47	& 0.39		& 	& 0.39	& 0.40	& 0.33	& 0.39	& 0.31	& 0.40	& 	& 0.43	& 0.46	& 0.43	& 0.26	& 	& 0.45  	\\
\bottomrule
\end{tabular*}
\caption{Q2. Mean absolute error of the SVM with all features and the mean baseline for all quality dimensions on all test sets, separated for training on the ground-truth scores of expert \#1, \#2, or \#3, or the mean scores (as in Table~\ref{table-features}) respectively. The value with the highest significance in each column is bold.}
\label{table-subjectiveness}
\end{table*}

%% file: table-human-machine.tex
\begin{table*}[t!]
\small
\renewcommand{\arraystretch}{0.96}
\centering
\setlength{\tabcolsep}{1.7pt}
\begin{tabular*}{\linewidth}{ll@{}crrrr@{$\;\;\;$}crrrrrr@{$\;\;\;$}crrrr@{$\;\;\;$}cr}
\toprule										
%		&					&	& \multicolumn{4}{c}{\bf Logical quality}		&	& \multicolumn{6}{c}{\bf Rhetorical quality}					&	& \multicolumn{4}{c}{\bf Dialectical quality}			\\		
%									\cmidrule(l@{2pt}r@{2pt}){4-7}				\cmidrule(l@{2pt}r@{2pt}){9-14}							\cmidrule(l@{2pt}r@{2pt}){16-19}					 	
\bf 	& \bf ``Approach''			&	& \bf Cog	& \bf LAc	& \bf LRe	& \bf LSu		&	& \bf Eff	& \bf Cla	& \bf Cre	& \bf App	& \bf Emo	& \bf Arr	& 	& \bf Rea	& \bf GAc	& \bf GRe	& \bf GSu	& 	& \bf OvQ	\\			
\midrule
Humans		& Expert \#1		&	& 0.32	& 0.32	& \bf $^\ddagger$0.23	& 0.29		&	& 0.32	& \gr 0.34	& \bf $^\ddagger$0.19	& \gr 0.30	& \bf 0.18	& \bf 0.27	&	& $^\dagger$0.22	& $^\ddagger$0.27	& \gr 0.40	& \gr 0.29	&	& 0.27	\\
			& Expert \#2		&	& 0.28	& 0.38	& 0.33	& 0.31		&	& $^\ddagger$0.24	& 0.31	& 0.27	& \gr 0.46	& \gr 0.25	& \gr 0.35	&	& 0.30	& $^\dagger$0.29	& \gr 0.41	& \gr 0.24	&	& $^\dagger$0.26	\\
			& Expert \#3		&	& \bf $^\ddagger$0.17	& \bf $^\ddagger$0.22	& \bf $^\ddagger$0.23	& \bf $^\ddagger$0.15		&	& \bf $^\ddagger$0.16	& \bf \phantom{$^\dagger$}0.26	& \phantom{$^\dagger$}0.27	& \bf \phantom{$^\dagger$}0.26	& \gr \phantom{$^\dagger$}0.27	& \phantom{$^\dagger$}0.28	&	& \bf $^\ddagger$0.16	& \bf $^\ddagger$0.26	& \bf $^\ddagger$0.22	& \bf $^\ddagger$0.08	&	& \bf $^\ddagger$0.14	\\
\midrule
SVM			& All features		&	& 0.36	& 0.38	& 0.42	& 0.34		& 	& 0.36	& 0.33	& 0.29	& 0.28	& 0.22	& 0.33	&	& 0.34	& 0.40	& 0.39	& 0.21	&	& 0.36  	\\
%			& BERT			&	& ??		& ??		& ??		& ??			& 	& ??		& ??		& ??		& ??		& ??		& ??		& 	& ??		& ??		& ??		& ??		& 	& ??	\\
%\midrule
%\multicolumn{2}{l}{Krippendorff's $\alpha$}&& 0.44	& 0.46	& 0.47	& 0.44		&	& 0.45	& 0.35	& 0.37	& 0.36	& 0.29	& 0.39	& 	& 0.50	& 0.44	& 0.42	& 0.27	& 	& 0.51 \\
\bottomrule
\end{tabular*}
\caption{Q3. Mean absolute error of each expert and the SVM with all features, %and BERT 
for all quality dimensions on all test sets, based on majority scores. The best value for each dimension is bold. Gray expert values are worse than the features; significant gains over the features are marked with $\dagger$ ($p\!<\!.05$) and $\ddagger$ ($p\!< \!.01$).}%For interpretation, the inter-rater agreement is given.}
\label{table-human-machine}
\end{table*}

%% file: coling20-quality-assessment-sum.tex
\section{Conclusion}
\label{sec:conclusion}

In this focused study, we have systematically benchmarked how well argument quality can be assessed computationally on an existing corpus annotated for several quality dimensions. Modeling subjectiveness in terms of sentiment, pronoun usage, and similar seems useful on the debate portal arguments included in the corpus, at least for logical and dialectical dimensions. However, the limited corpus size naturally makes it hard to find more complex features that robustly predict argument quality. In addition, the correlation of quality and length in the corpus limits the generalizability of our findings. This calls for more large-scale and balanced argument quality corpora. First attempts in this direction have been \nopagebreak{made} \cite{toledo:2019}, but the comprehensive view on quality of the corpus used here has no equal so far.\,\,\,

\section*{Acknowledgments}

We thank Eyke H\"ullermeier for suggestions on the experimental setup, Gabriela Molina Le\'{o}n for feedback on early drafts, and the anonymous reviewers for their helpful comments.

% NOTE. Needed to make the conclusion fit on page 4
%\pagebreak

%% file: coling20-quality-assessment-frame.bbl
\begin{thebibliography}{}

\bibitem[\protect\citename{Aggarwal and Zhai}2012]{aggarwal:2012}
Charu~C. Aggarwal and ChengXiang Zhai, 2012.
\newblock {\em A Survey of Text Classification Algorithms}, pages 163--222.
\newblock Springer US, Boston, MA.

\bibitem[\protect\citename{Chang and Lin}2011]{chang:2011}
Chih-Chung Chang and Chih-Jen Lin.
\newblock 2011.
\newblock {LIBSVM}: {A} library for support vector machines.
\newblock {\em ACM Transactions on Intelligent Systems and Technology},
  2:27:1--27:27.

\bibitem[\protect\citename{Devlin \bgroup et al.\egroup }2018]{devlin:2018}
Jacob Devlin, Ming-Wei Chang, Kenton Lee, and Kristina Toutanova.
\newblock 2018.
\newblock {BERT}: {P}re-training of deep bidirectional transformers for
  language understanding.
\newblock {\em arXiv preprint arXiv:1810.04805}.

\bibitem[\protect\citename{El~Baff \bgroup et al.\egroup }2018]{elbaff:2018}
Roxanne El~Baff, Henning Wachsmuth, Khalid Al-Khatib, and Benno Stein.
\newblock 2018.
\newblock Challenge or empower: {R}evisiting argumentation quality in a news
  editorial corpus.
\newblock In {\em Proceedings of the 22nd Conference on Computational Natural
  Language Learning}, pages 454--464, Brussels, Belgium, October. Association
  for Computational Linguistics.

\bibitem[\protect\citename{El~Baff \bgroup et al.\egroup }2020]{elbaff:2020a}
Roxanne El~Baff, Henning Wachsmuth, Khalid Al~Khatib, and Benno Stein.
\newblock 2020.
\newblock Analyzing the persuasive effect of style in news editorial
  argumentation.
\newblock In {\em Proceedings of the 58th Annual Meeting of the Association for
  Computational Linguistics}, pages 3154--3160, Online, July. Association for
  Computational Linguistics.

\bibitem[\protect\citename{Gleize \bgroup et al.\egroup }2019]{gleize:2019}
Martin Gleize, Eyal Shnarch, Leshem Choshen, Lena Dankin, Guy Moshkowich, Ranit
  Aharonov, and Noam Slonim.
\newblock 2019.
\newblock Are you convinced? {C}hoosing the more convincing evidence with a
  {S}iamese network.
\newblock In {\em Proceedings of the 57th Annual Meeting of the Association for
  Computational Linguistics}, pages 967--976, Florence, Italy, July.
  Association for Computational Linguistics.

\bibitem[\protect\citename{Gretz \bgroup et al.\egroup }2020]{gretz:2020}
Shai Gretz, Roni Friedman, Edo Cohen-Karlik, Assaf Toledo, Dan Lahav, Ranit
  Aharonov, and Noam Slonim.
\newblock 2020.
\newblock A large-scale dataset for argument quality ranking: {C}onstruction
  and analysis.
\newblock In {\em Proceedings of the Thirty-Fourth AAAI Conference on
  Artificial Intelligence}, pages 7805--7813. AAAI.

\bibitem[\protect\citename{Habernal and Gurevych}2016a]{habernal:2016b}
Ivan Habernal and Iryna Gurevych.
\newblock 2016a.
\newblock What makes a convincing argument? {E}mpirical analysis and detecting
  attributes of convincingness in web argumentation.
\newblock In {\em Proceedings of the 2016 Conference on Empirical Methods in
  Natural Language Processing}, pages 1214--1223. Association for Computational
  Linguistics.

\bibitem[\protect\citename{Habernal and Gurevych}2016b]{habernal:2016a}
Ivan Habernal and Iryna Gurevych.
\newblock 2016b.
\newblock Which argument is more convincing? {A}nalyzing and predicting
  convincingness of web arguments using bidirectional lstm.
\newblock In {\em Proceedings of the 54th Annual Meeting of the Association for
  Computational Linguistics (Volume 1: Long Papers)}, pages 1589--1599.
  Association for Computational Linguistics.

\bibitem[\protect\citename{Johnson and Blair}2006]{johnson:2006}
Ralph~H. Johnson and J.~Anthony Blair.
\newblock 2006.
\newblock {\em Logical Self-defense}.
\newblock Intern.\ Debate Education Association.

\bibitem[\protect\citename{Ke and Ng}2019]{ke:2019}
Zixuan Ke and Vincent Ng.
\newblock 2019.
\newblock Automated essay scoring: {A} survey of the state of the art.
\newblock In {\em Proceedings of the Twenty-Eighth International Joint
  Conference on Artificial Intelligence}, pages 6300--6308. IJCAI.

\bibitem[\protect\citename{Lukin \bgroup et al.\egroup }2017]{lukin:2017}
Stephanie Lukin, Pranav Anand, Marilyn Walker, and Steve Whittaker.
\newblock 2017.
\newblock {Argument Strength is in the Eye of the Beholder: {A}udience Effects
  in Persuasion}.
\newblock In {\em Proceedings of the 15th Conference of the European Chapter of
  the Association for Computational Linguistics: Volume 1, Long Papers}, pages
  742--753. Association for Computational Linguistics.

\bibitem[\protect\citename{Mikolov \bgroup et al.\egroup }2018]{mikolov:2018}
Tomas Mikolov, Edouard Grave, Piotr Bojanowski, Christian Puhrsch, and Armand
  Joulin.
\newblock 2018.
\newblock Advances in pre-training distributed word representations.
\newblock In {\em Proceedings of the Eleventh International Conference on
  Language Resources and Evaluation ({LREC} 2018)}, Miyazaki, Japan, May.
  European Language Resources Association (ELRA).

\bibitem[\protect\citename{Persing and Ng}2015]{persing:2015}
Isaac Persing and Vincent Ng.
\newblock 2015.
\newblock Modeling argument strength in student essays.
\newblock In {\em Proceedings of the 53rd Annual Meeting of the Association for
  Computational Linguistics and the 7th International Joint Conference on
  Natural Language Processing (Volume 1: Long Papers)}, pages 543--552.
  Association for Computational Linguistics.

\bibitem[\protect\citename{Persing \bgroup et al.\egroup }2010]{persing:2010}
Isaac Persing, Alan Davis, and Vincent Ng.
\newblock 2010.
\newblock Modeling organization in student essays.
\newblock In {\em Proceedings of the 2010 Conference on Empirical Methods in
  Natural Language Processing}, pages 229--239. Association for Computational
  Linguistics.

\bibitem[\protect\citename{Potash \bgroup et al.\egroup }2017]{potash:2017}
Peter Potash, Robin Bhattacharya, and Anna Rumshisky.
\newblock 2017.
\newblock Length, interchangeability, and external knowledge: {O}bservations
  from predicting argument convincingness.
\newblock In {\em Proceedings of the Eighth International Joint Conference on
  Natural Language Processing (Volume 1: Long Papers)}, pages 342--351, Taipei,
  Taiwan, November. Asian Federation of Natural Language Processing.

\bibitem[\protect\citename{Potthast \bgroup et al.\egroup }2019]{potthast:2019}
Martin Potthast, Lukas Gienapp, Florian Euchner, Nick Heilenk{\"o}tter, Nico
  Weidmann, Henning Wachsmuth, Benno Stein, and Matthias Hagen.
\newblock 2019.
\newblock Argument search: {A}ssessing argument relevance.
\newblock In {\em 42nd International ACM Conference on Research and Development
  in Information Retrieval (SIGIR 2019)}, pages 1117--1120. ACM, July.

\bibitem[\protect\citename{Rittman \bgroup et al.\egroup }2004]{rittman:2004}
Robert Rittman, Nina Wacholder, Paul Kantor, Kwong~Bor Ng, Tomek StrzaIkowski,
  and Ying Sun.
\newblock 2004.
\newblock Adjectives as indicators of subjectivity in documents.
\newblock In {\em Proceedings of the 67th ASIS\&T Annual Meeting}, pages
  349--359. American Society for Information Science.

\bibitem[\protect\citename{Stab and Gurevych}2017]{stab:2017}
Christian Stab and Iryna Gurevych.
\newblock 2017.
\newblock Recognizing insufficiently supported arguments in argumentative
  essays.
\newblock In {\em Proceedings of the 15th Conference of the {E}uropean Chapter
  of the Association for Computational Linguistics: Volume 1, Long Papers},
  pages 980--990. Association for Computational Linguistics.

\bibitem[\protect\citename{Stamatatos}2009]{stamatatos:2009}
Efstathios Stamatatos.
\newblock 2009.
\newblock A survey of modern authorship attribution methods.
\newblock {\em Journal of the American Society for Information Science and
  Technology}, 60(3):538--556.

\bibitem[\protect\citename{Stede and Schneider}2018]{stede:2018}
Manfred Stede and Jodi Schneider.
\newblock 2018.
\newblock {\em Argumentation Mining}.
\newblock Number~40 in Synthesis Lectures on Human Language Technologies.
  Morgan \& Claypool.

\bibitem[\protect\citename{Toledo \bgroup et al.\egroup }2019]{toledo:2019}
Assaf Toledo, Shai Gretz, Edo Cohen-Karlik, Roni Friedman, Elad Venezian, Dan
  Lahav, Michal Jacovi, Ranit Aharonov, and Noam Slonim.
\newblock 2019.
\newblock Automatic argument quality assessment - {N}ew datasets and methods.
\newblock In {\em Proceedings of the 2019 Conference on Empirical Methods in
  Natural Language Processing and the 9th International Joint Conference on
  Natural Language Processing (EMNLP-IJCNLP)}, pages 5625--5635. Association
  for Computational Linguistics.

\bibitem[\protect\citename{van Eemeren and Grootendorst}2004]{vaneemeren:2004}
Frans~H. van Eemeren and Rob Grootendorst.
\newblock 2004.
\newblock {\em A Systematic Theory of Argumentation: {T}he Pragma-Dialectical
  Approach}.
\newblock Cambridge University Press, Cambridge, UK.

\bibitem[\protect\citename{Wachsmuth \bgroup et al.\egroup
  }2016]{wachsmuth:2016}
Henning Wachsmuth, Khalid Al~Khatib, and Benno Stein.
\newblock 2016.
\newblock Using argument mining to assess the argumentation quality of essays.
\newblock In {\em Proceedings of COLING 2016, the 26th International Conference
  on Computational Linguistics: Technical Papers}, pages 1680--1691. The COLING
  2016 Organizing Committee.

\bibitem[\protect\citename{Wachsmuth \bgroup et al.\egroup
  }2017a]{wachsmuth:2017d}
Henning Wachsmuth, Nona Naderi, Ivan Habernal, Yufang Hou, Graeme Hirst, Iryna
  Gurevych, and Benno Stein.
\newblock 2017a.
\newblock Argumentation quality assessment: {T}heory vs. practice.
\newblock In {\em Proceedings of the 55th Annual Meeting of the Association for
  Computational Linguistics (Volume 2: Short Papers)}, pages 250--255.
  Association for Computational Linguistics.

\bibitem[\protect\citename{Wachsmuth \bgroup et al.\egroup
  }2017b]{wachsmuth:2017b}
Henning Wachsmuth, Nona Naderi, Yufang Hou, Yonatan Bilu, Vinodkumar
  Prabhakaran, Alberdingk~Tim Thijm, Graeme Hirst, and Benno Stein.
\newblock 2017b.
\newblock Computational argumentation quality assessment in natural language.
\newblock In {\em Proceedings of the 15th Conference of the European Chapter of
  the Association for Computational Linguistics: Volume 1, Long Papers}, pages
  176--187. Association for Computational Linguistics.

\bibitem[\protect\citename{Wachsmuth \bgroup et al.\egroup
  }2017c]{wachsmuth:2017a}
Henning Wachsmuth, Benno Stein, and Yamen Ajjour.
\newblock 2017c.
\newblock ``{P}age{R}ank'' for argument relevance.
\newblock In {\em Proceedings of the 15th Conference of the European Chapter of
  the Association for Computational Linguistics: Volume 1, Long Papers}, pages
  1117--1127. Association for Computational Linguistics.

\bibitem[\protect\citename{Walton}2006]{walton:2006}
Douglas Walton.
\newblock 2006.
\newblock {\em Fundamentals of Critical Argumentation}.
\newblock Cambridge University Press.

\bibitem[\protect\citename{Wei \bgroup et al.\egroup }2016]{wei:2016}
Zhongyu Wei, Yang Liu, and Yi~Li.
\newblock 2016.
\newblock Is this post persuasive? {R}anking argumentative comments in online
  forum.
\newblock In {\em Proceedings of the 54th Annual Meeting of the Association for
  Computational Linguistics (Volume 2: Short Papers)}, pages 195--200.
  Association for Computational Linguistics.

\bibitem[\protect\citename{Yang \bgroup et al.\egroup }2019]{yang:2019}
Wonsuk Yang, Seungwon Yoon, Ada Carpenter, and Jong Park.
\newblock 2019.
\newblock Nonsense!: {Q}uality control via two-step reason selection for
  annotating local acceptability and related attributes in news editorials.
\newblock In {\em Proceedings of the 2019 Conference on Empirical Methods in
  Natural Language Processing and the 9th International Joint Conference on
  Natural Language Processing (EMNLP-IJCNLP)}, pages 2954--2963. Association
  for Computational Linguistics.

\end{thebibliography}
